\documentclass{article}
\usepackage[utf8]{inputenc}
\usepackage{charter} 
\usepackage{geometry}
\usepackage{authblk}
\usepackage{siunitx}
\usepackage{amsfonts}
\usepackage{chemformula}
\usepackage{hyperref}
\usepackage{graphicx}
\usepackage{caption}
\usepackage{subcaption}
\usepackage{listings}
\usepackage[title]{appendix}
\usepackage[style=nature]{biblatex}
\usepackage{xcolor}

\usepackage{xparse}

\NewDocumentCommand \G {e_}{%
  \IfNoValueTF{#1}{%
    \mathcal{G}
  }{%
    \mathcal{G}_{\mathrm{#1}}
  }%
}

\NewDocumentCommand \LG {e_}{%
  \IfNoValueTF{#1}{%
    \mathcal{L(G)}
  }{%
    \mathcal{L(G_{\mathrm{#1}})}
  }%
}

\NewDocumentCommand \LpG {e_}{%
  \IfNoValueTF{#1}{%
    \mathcal{L'(G)}
  }{%
    \mathcal{L'(G_{\mathrm{#1}})}
  }%
}

\renewcommand{\vec}[1]{\mathbf{#1}}

\title{Efficient, Interpretable Graph Neural Network Representation for Angle-dependent Properties and its Application to Optical Spectroscopy}

\author[1]{Tim Hsu\thanks{Corresponding Author, hsu16@llnl.gov}}
\author[2]{Tuan Anh Pham\thanks{Corresponding Author, pham16@llnl.gov}}
\author[2]{Nathan Keilbart}
\author[2]{Stephen Weitzner}
\author[2]{James Chapman}
\author[3]{Penghao Xiao}
\author[2]{S. Roger Qiu}
\author[1]{Xiao Chen}
\author[2]{Brandon C. Wood\thanks{Corresponding Author, wood37@llnl.gov}}

\affil[1]{Center for Applied Scientific Computing, Lawrence Livermore National Laboratory, Livermore, CA, USA}
\affil[2]{Materials Science Division, Lawrence Livermore National Laboratory, Livermore, CA, USA}
\affil[3]{Department of Physics and Atmospheric Science, Dalhousie University, Halifax, NS, Canada}

\date{}

\begin{document}

\maketitle

\begin{abstract}
Graph neural networks are attractive for learning properties of atomic structures thanks to their intuitive graph encoding of atoms and bonds. However, conventional encoding does not include angular information, which is critical for describing atomic arrangements in disordered systems. In this work, we extend the recently proposed ALIGNN encoding, which incorporates bond angles, to also include dihedral angles (ALIGNN-$d$). This simple extension leads to a memory-efficient graph representation that captures the complete geometry of atomic structures. ALIGNN-$d$ is applied to predict the infrared optical response of dynamically disordered Cu(II) aqua complexes, leveraging the intrinsic interpretability to elucidate the relative contributions of individual structural components. Bond and dihedral angles are found to be critical contributors to the fine structure of the absorption response, with distortions representing transitions between more common geometries exhibiting the strongest absorption intensity. Future directions for further development of ALIGNN-$d$ are discussed.
\end{abstract}
\section{Introduction}
In materials science, graph neural networks (GNNs) have gained popularity as a surrogate model for learning properties of materials and molecular systems \cite{gilmer2017neural, coley2017convolutional, schutt2018schnet, xie2018crystal, yang2019analyzing, chen2019}. This popularity is partly due to the intuitive, physically informed graph encoding that represents atoms with nodes and bonds with edges. However, beyond atom and bond features, encoding further structural information can be helpful or even required for accurate prediction of certain properties. For example, the bond angle information is necessary for correctly capturing electronic structure and bond hybridization \cite{linker2020, X-angle1, X-angle2, X-angle3}. Likewise, in machine learning potentials where accurate energy prediction is needed, three-body (bond angle) and higher-order terms are usually included in the descriptor \cite{behler2007generalized, samanta2018representing, lindsey2017chimes}.

One practical scenario in which three- (bond angle) and four-body (dihedral angle) interactions can be critical is spectroscopy prediction. These interactions alter the local electronic structure in ways that are often detectable in X-ray and optical absorption experiments, as well as in local chemical probes such as nuclear magnetic resonance. Indeed, the power of these experimental techniques draws in part from their sensitivity to local geometric and electronic environments; however, spectral features are often convoluted and not always straightforward to predict. This is particularly true for disordered and distorted atomic environments, which are common to interfaces and glassy systems, where slight perturbations in geometry can greatly impact resulting properties.~\cite{xps,water,photo,solvation} 

\begin{figure}
    \centering
    \includegraphics[width=0.6\textwidth]{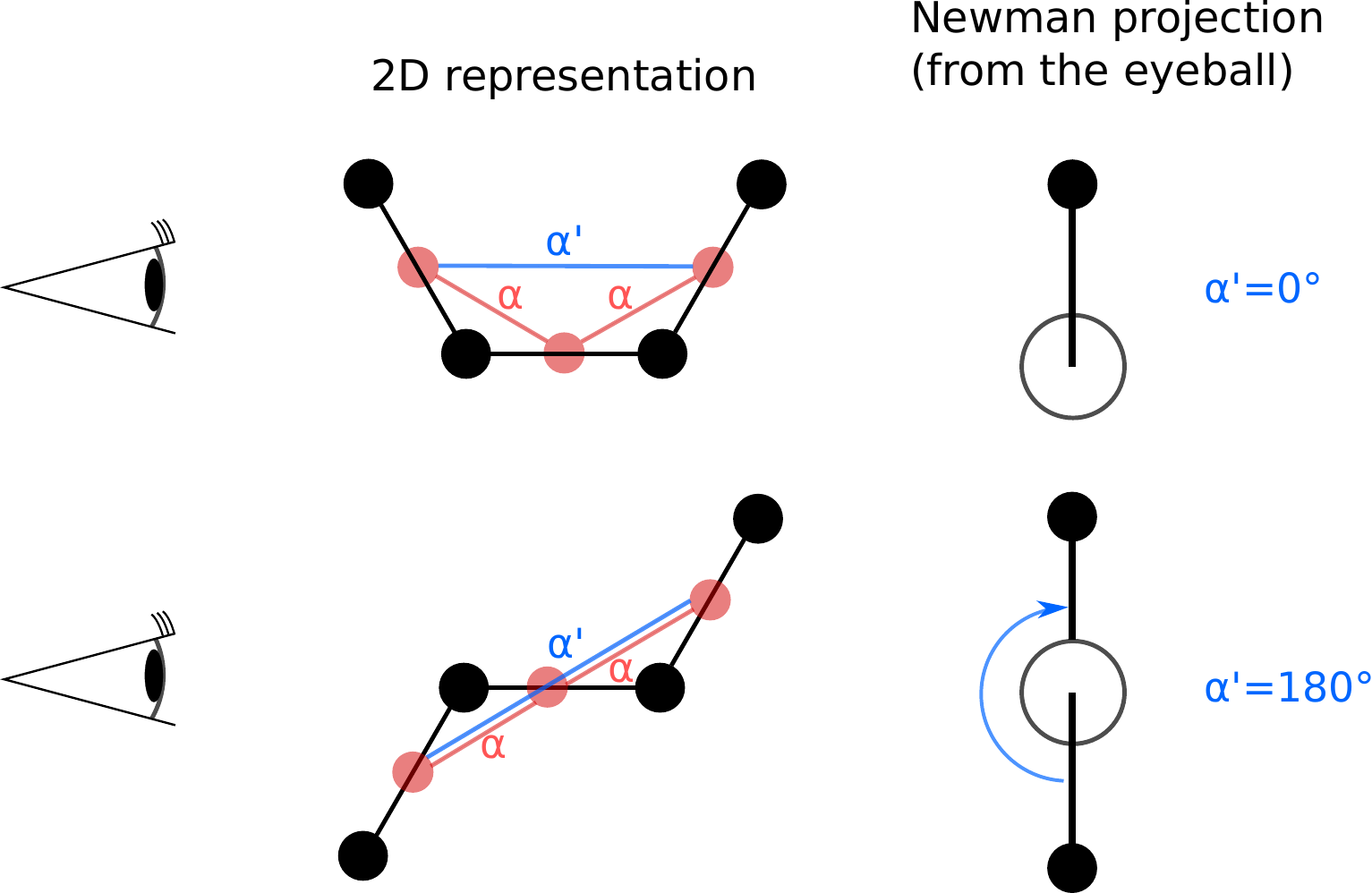}
    \caption{The line graphs, shown as blue and red nodes and edges, encode angular information absent in the original atomic graph (black). Red edges capture bond angles $\alpha$, and blue edges capture dihedral angles $\alpha'$ (defined as the clockwise angle in the Newman projection between two bonds sharing a common bond). Distinguishing these two configurations requires inclusion of the dihedral angle in the graph encoding.}
    \label{fig:alignnd-demo}
\end{figure}

Unfortunately, conventional GNNs for atomic systems do not encode angular information. Recently, several approaches have been proposed to explicitly encode bond angles \cite{park2020developing, klicpera2020directional, decost2021atomistic} or directional information from which bond angles can be implicitly retrieved \cite{schutt2021equivariant, CHAPMAN2020109483}.
Many of these were designed for non-periodic molecular structures. Alternatively, the ALIGNN approach \cite{decost2021atomistic}, which explicitly represents bond angles (three-body terms) as edges of line graphs, is a general formulation applicable to both non-periodic molecular graphs and periodic crystal graphs \cite{xie2018crystal, chapman2021sgop}. However, despite its advantages, the ALIGNN encoding may not capture the full structural information of a local geometric environment. This limitation is demonstrated by the example in Fig.~\ref{fig:alignnd-demo}.

In this work, we expand the ALIGNN encoding to include dihedral angle information. This enhanced graph representation, named ALIGNN-\textit{d}, provides more complete structural information and greater interpretability. To demonstrate these advantages, we train GNN models based on ALIGNN-\textit{d} alongside competing graph representations to predict infrared optical absorption spectral signatures of Cu(II) aqua complexes. These complexes are optically active and known to have high absorption sensitivity to local geometry. Utilizing configurations derived from first-principles molecular dynamics simulations, we specifically probe the role of local distortion in the GNN encoding and resulting spectroscopic signatures. Based on the results, we identify three primary advantages of the ALIGNN-\textit{d} representation. First, ALIGNN-\textit{d} is a compact description that leads to roughly the same predictive accuracy as the maximally connected graph (in which all pairwise bonds are encoded) but with greater memory efficiency. Second, the loss convergence based on the auxiliary line-graph encoding (ALIGNN or ALIGNN-\textit{d}) is faster and more stable than alternatives without line graphs. Third, ALIGNN-\textit{d} enables an intuitive approach to model interpretability thanks to the explicit graph representation of bond and dihedral angles.

\section{Results}

\subsection{Optical response of Cu(II) aqua complexes}
The capabilities of ALIGNN-$d$ were demonstrated by predicting infrared optical absorption spectral signatures of Cu(II) aqua complexes. These systems are broadly representative of transition metal molecular complexes that absorb optically due to \textit{d}–\textit{d} transitions, which are leveraged in a variety of materials applications and biological processes \cite{Qiu2015}. Configurations were obtained from \textit{ab-initio} molecular dynamics simulations (AIMD), and optical transitions were calculated using time-dependent density functional theory, as described in the Methods section. 

\begin{figure}
    \centering
    \begin{subfigure}[b]{0.24\textwidth}
        \centering
        \includegraphics[width=\textwidth]{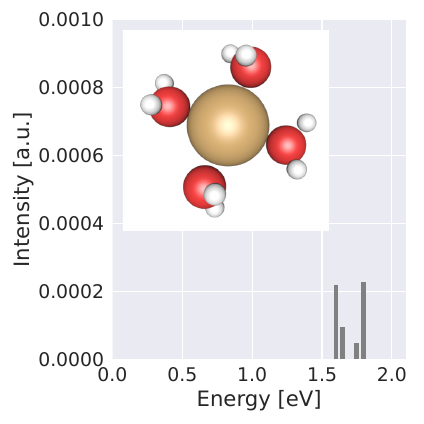}
        \caption{}
    \end{subfigure}
    \hfill
    \begin{subfigure}[b]{0.24\textwidth}
        \centering
        \includegraphics[width=\textwidth]{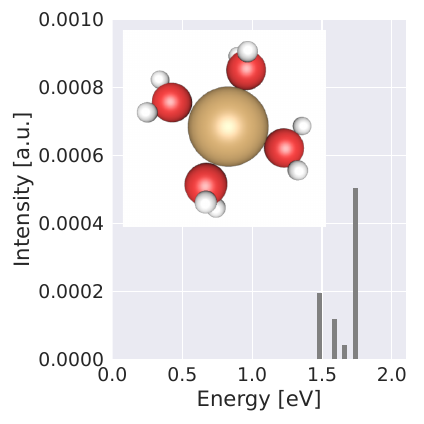}
        \caption{}
    \end{subfigure}
    \hfill
    \begin{subfigure}[b]{0.24\textwidth}
        \centering
        \includegraphics[width=\textwidth]{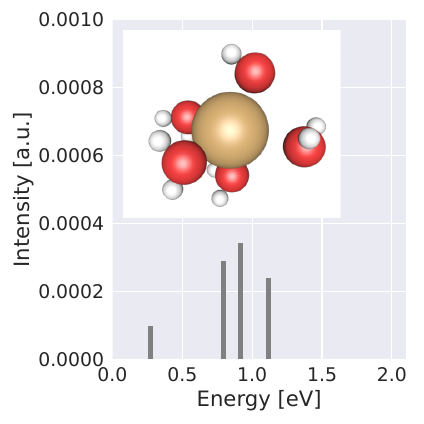}
        \caption{}
    \end{subfigure}
    \hfill
    \begin{subfigure}[b]{0.24\textwidth}
        \centering
        \includegraphics[width=\textwidth]{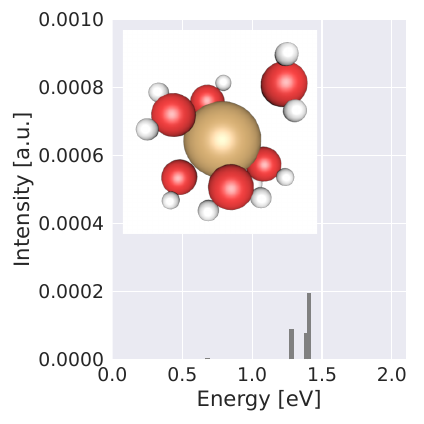}
        \caption{}
    \end{subfigure}
    \caption{Structures of the Cu(II) aqua complex from AIMD (figure insets) and their corresponding simulated optical transitions. The central copper ion is surrounded by either (a, b) four, (c) five, or (d) six water molecules, with fivefold coordination being the most common. Slight structural perturbations (a, b) and changes in coordination (c, d) result in noticeably different shifts of the energies and intensities. The discrete spectral peaks were computed from time-dependent density functional theory (TDDFT) simulations.}
    \label{fig:clusters-and-peaks}
\end{figure}

Beyond the practical implications, the high sensitivity of optical properties to the local geometry of the Cu(II) aqua complexes provides an excellent test of ALIGNN-$d$. This can be seen in Fig.~\ref{fig:clusters-and-peaks}, which plots the optical transitions in the infrared regime, computed from time-dependent density functional theory (TDDFT), for complexes with different instantaneous coordination numbers. In these complexes, the coordination number is found to fluctuate between four and six, with fivefold coordination as the most common and sixfold coordination as the least common \cite{Qiu2015, Pasquarello2001}. The results clearly indicate that the infrared optical absorption is highly sensitive to the water coordination number, with little similarity between the spectral response of the sampled configurations. Moreover, complexes with the same coordination number and visibly similar atomic configurations (a, b) can generate infrared absorption profiles with noticeably different peak locations and intensities, confirming that absorption in this frequency regime is also sensitive to subtle differences in the local bonding character. The physical origin of this behavior is connected to the fundamental nature of the \textit{d}–\textit{d} transitions, which are nominally symmetry forbidden in ideal structures but are activated by thermal distortions. We utilize these distortions and their spectral response as our basis to test the benefits of ALIGNN-$d$. Further analysis of the correlation between geometry/shape information and spectral response can be found in Supplementary Information (Fig.~S1).

\subsection{Graph representations and prediction accuracy}
\begin{figure}
    \centering
    \includegraphics[width=0.7\textwidth]{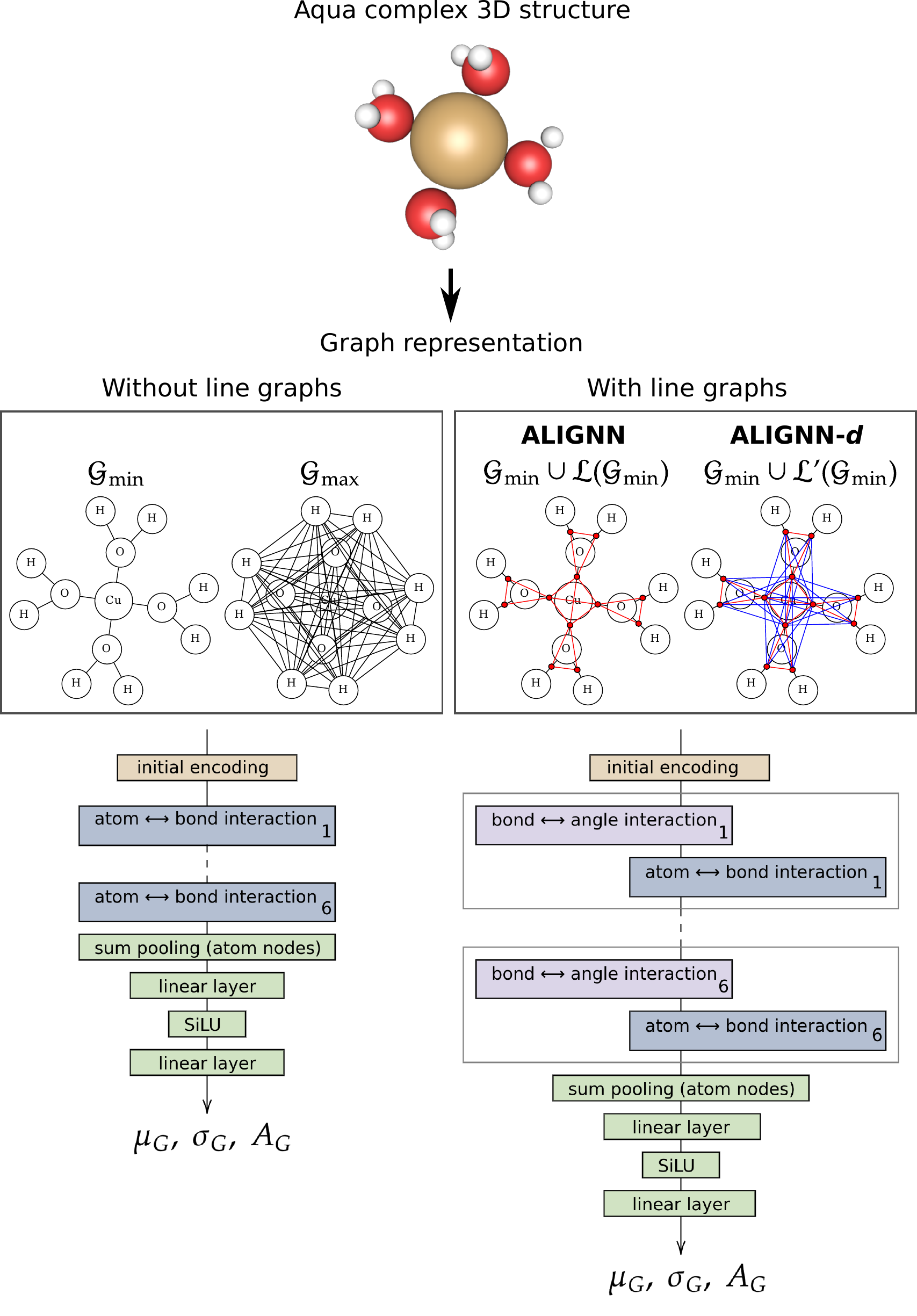}
    \caption{The structure-to-spectrum flow consists of conversion to graph representation, followed by spectroscopy prediction from GNN layers. The four different graph representations discussed in this work are shown schematically. The final output is a Gaussian curve parameterized by mean $\mu_G$, standard deviation $\sigma_G$, and amplitude $A_G$, approximating the TDDFT-calculated discrete spectral peaks.}
    \label{fig:flow}
\end{figure}

We first introduce our workflow for encoding the molecular features and predicting the spectroscopic signatures. Summarised in Fig.~\ref{fig:flow}, this involves converting the atomic structure of the Cu(II) aqua complexes into a graph representation, followed by predicting the key spectral features from GNNs. The specific outputs of this procedure are unnormalized Gaussian functions that approximate the absorption spectra of the complexes, parameterized by the mean peak position $\mu_G$, spectral width $\sigma_G$, and intensity $A_G$.

As a key component of the workflow, we compare four graph representations for encoding the molecular structures. First, we consider the minimally connected graph $\G_{min}$ that encodes only the minimal number of edges to connect the nearest-neighbor atomic bonds. In this regard, $\G_{min}$ contains the least amount of structural information, as the bond angles and dihedral angles are not implicitly included. Second, as the opposite extreme, we consider the maximally connected graph $\G_{max}$, which represents the brute-force approach that encodes all the pairwise bonds, thereby yielding complete geometric information \cite{klicpera2020directional}. Third, following the ALIGNN formulation, we add bond angle information to $\G_{min}$ by adding the corresponding line graph $\LG$. Finally, we extend the ALIGNN formulation to explicitly represent both bond and dihedral angles in the line graph encoding, which we denote as $\LpG$. 

We emphasize that ALIGNN and ALIGNN$-d$, written as the union sets $\G_{min} \cup \LG_{min}$ and $\G_{min} \cup \LG_{min}$, are expected to have improved representation power with respect to $\G_{min}$. In fact, it is known that atomic numbers, bond lengths, bond angles, and dihedral angles together can fully describe the complete structure of a molecular system. This follows the principle of the Z-matrix \cite{parsons2005practical}, which has been shown to uniquely convert this set of quantities back to the exact Cartesian coordinates of the atoms. Whereas ALIGNN encodes atom, bond, and bond angle features, the addition of dihedral angle information (i.e., four-body terms) in ALIGNN$-d$ completes the Z-matrix and is therefore capable of fully describing the atomic structure. As a result, any complex geometric feature (including distortions, chirality, and disordered configurations) can be exactly represented without explicitly including higher-order terms. In other words, ALIGNN$-d$ implicitly has the same representation power as $\G_{max}$, despite its considerably smaller basis.

These four graph representations were applied to encode the dynamically fluctuating geometries of the Cu(II) aqua complex. The number of edges in each representation is listed in Table~\ref{tab:num-edges} for instantaneous configurations with different water coordination numbers. Although constructing line graphs on top of $\G_{min}$ introduces additional edges, the total number is always less than for $\G_{max}$. In particular, $\G_{min} \cup \LpG_{min}$ has roughly 31--35\% fewer edges, which translates to more efficient memory usage. We further note that inclusion of line graphs does not introduce additional nodes, since the nodes of $\LG$ are identical to the edges of $\G$.

\begin{table}[h]
    \centering
    \caption{Number of edges in each graph representation}
    \begin{tabular}{l | c c c c}
     & $\G_{min}$ & $\G_{min} \cup \LG_{min}$, ALIGNN & $\G_{min} \cup \LpG_{min}$, ALIGNN-$d$ & $\G_{max}$ \\
    \hline
    4-coordinated &   12 &   30  &   54 &  78 \\
    5-coordinated &   15 &   40  &   80 & 120 \\
    6-coordinated &   18 &   51  &  111 & 171 \\
    \end{tabular}
    \label{tab:num-edges}
\end{table}

\begin{figure}
    \centering
    \begin{subfigure}[b]{0.32\textwidth}
        \centering
        \includegraphics[width=\textwidth]{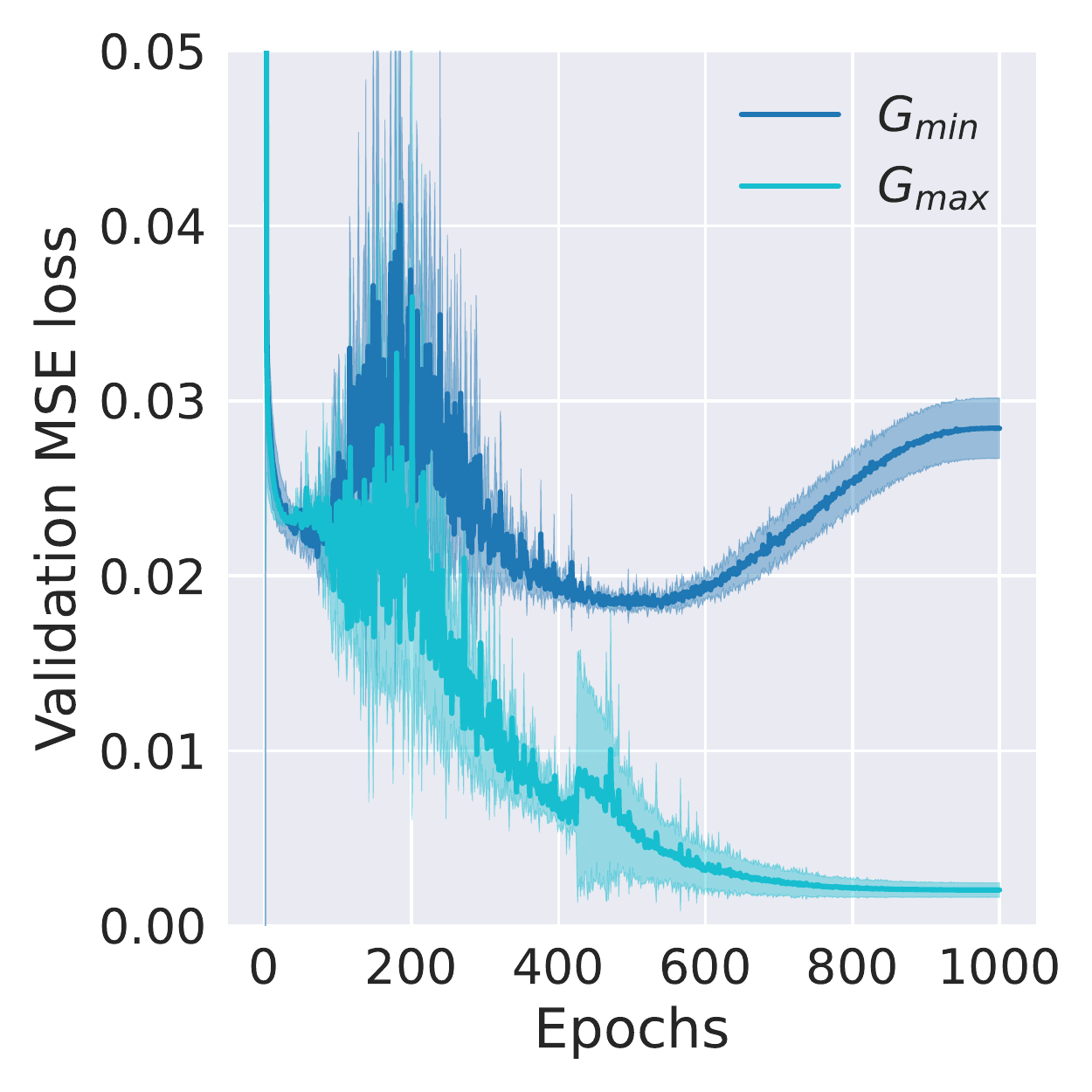}
        \caption{Without line graphs}
    \end{subfigure}
    \hfill
    \begin{subfigure}[b]{0.32\textwidth}
        \centering
        \includegraphics[width=\textwidth]{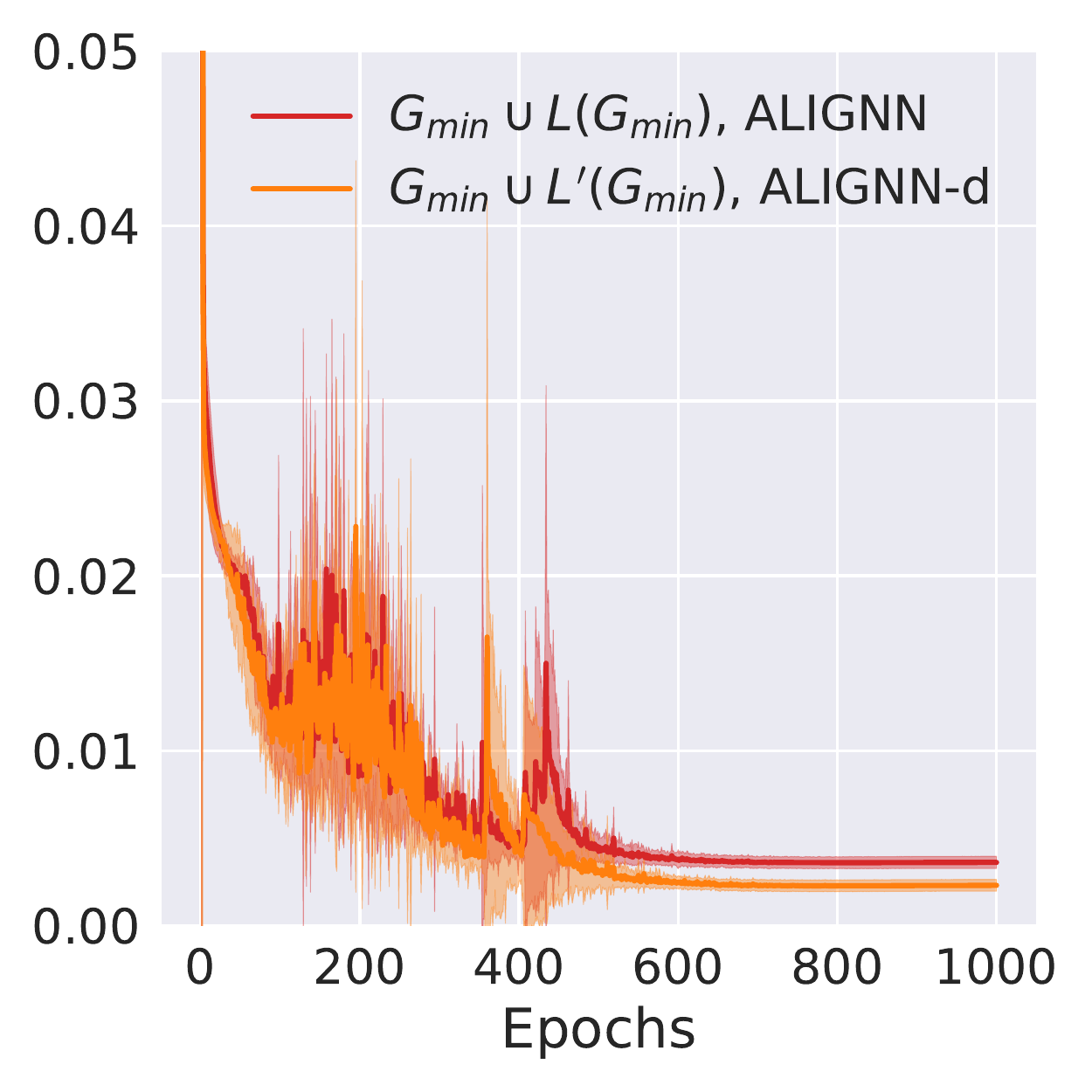}
        \caption{With line graphs}
    \end{subfigure}
    \hfill
    \begin{subfigure}[b]{0.32\textwidth}
        \centering
        \includegraphics[width=\textwidth]{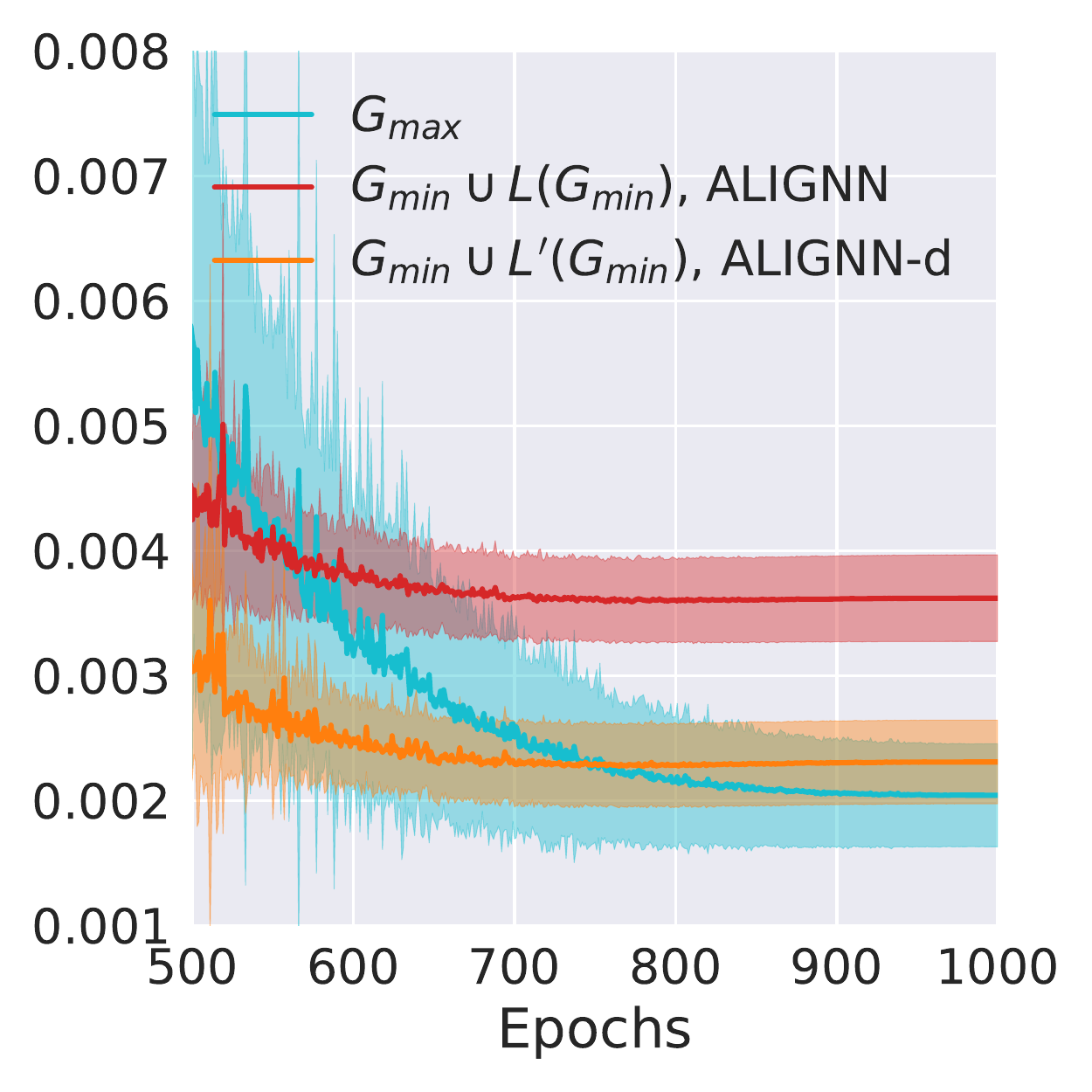}
        \caption{Combined and magnified}
    \end{subfigure}
    \caption{Validation losses during training for the four graph representations: (a) the representations without line graphs are $\G_{min}$ and $\G_{max}$; and (b) the representations with line graphs are $\G_{min} \cup \LG_{min}$ and $\G_{min} \cup \LpG_{min}$. Panel (c) combines (a) and (b) over a magnified region. Each training was repeated eight times, with standard deviations shown as semi-transparent regions and averages shown as solid lines.}
    \label{fig:val-loss}
\end{figure}

We proceed to explicitly demonstrate the performance of the four graph representations by comparing their corresponding expressive power. This translates to model accuracy, as measured via validation losses during training. As expected, our results, shown in Fig.~\ref{fig:val-loss}, indicate that the inclusion of auxiliary line graphs improves model performance. For instance, the use of $\LG$ and $\LpG$ leads to significant improvement over the minimum baseline $\G_{min}$, which is prone to overfitting (Fig.~\ref{fig:val-loss}a). Similarly, the losses based on representations with auxiliary line graphs converge faster with respective to the number of epochs, and are more stable than those without line graphs (Fig.~\ref{fig:val-loss}b).

Most importantly, we find that the inclusion of dihedral angle encoding in $\LpG_{min}$ leads to noticeably better performance over $\LG_{min}$. Without dihedral angles, $\G_{min} \cup \LG_{min}$, or ALIGNN, is limited in its expressive power compared to the maximally connected graph $\G_{max}$. On the other hand, the performance of $\G_{min} \cup \LpG_{min}$, or ALIGNN-$d$, is about the same as $\G_{max}$, closing the error gap between ALIGNN and $\G_{max}$ (Fig.~\ref{fig:val-loss}c). This verifies that ALIGNN$-d$ fully describes the atomic structure. 

\begin{figure}
    \centering
    \centering
    \includegraphics[width=\textwidth]{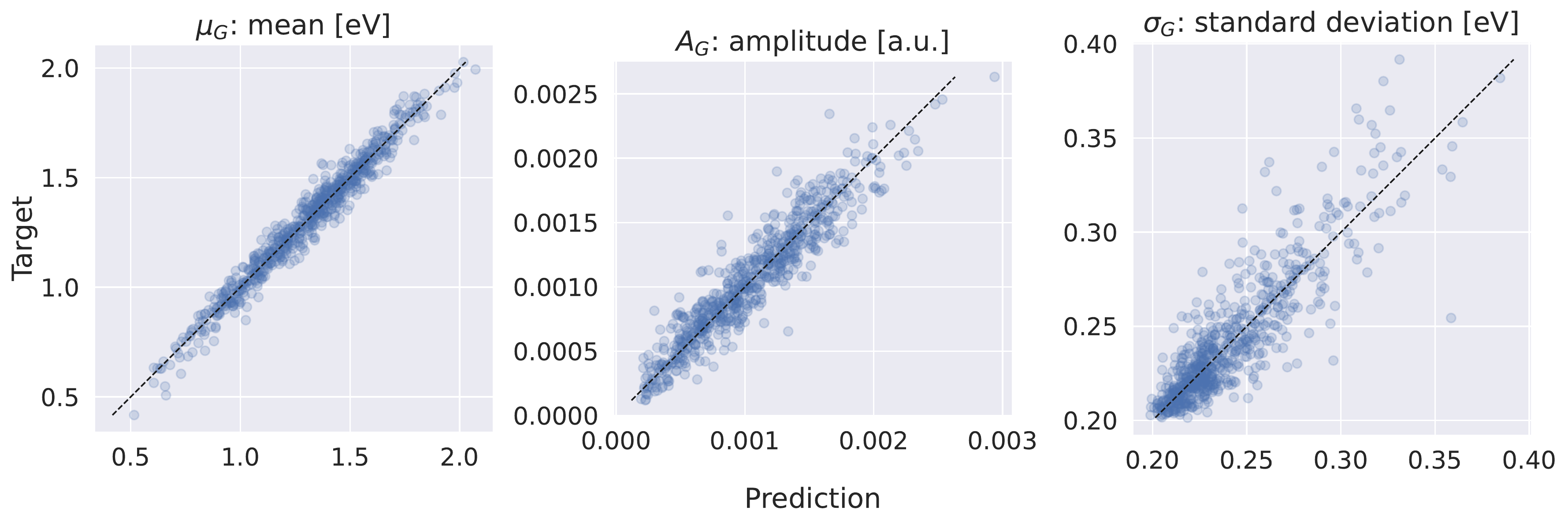}
    \caption{Parity plots of the predicted and explicitly computed optical absorption responses of different instantaneous configurations of the Cu(II) aqua complex based on the ALIGNN-$d$ encoding. The predicted and target Gaussian functions are parameterized by mean $\mu_G$, amplitude $A_G$, and standard deviation $\sigma_G$.}
    \label{fig:model-acc}
\end{figure}

Finally, we validate the accuracy of the ALIGNN-$d$ approach in predicting the infrared optical absorption spectra of the Cu(II) aqua complexes. Results presented in Fig.~\ref{fig:model-acc} show that our model provides accurate prediction of the mean $\mu_G$ and amplitude $A_G$ of the optical spectra, while yielding reasonable results for the standard deviation $\sigma_G$. In addition, we include in Supplementary Information a set of randomly sampled predicted peaks versus their target peaks (Fig.~S3), from which it is clear that ALIGNN-$d$ is capable of accurately predicting a wide variety of optical absorption spectral signatures.  

\section{Discussion}

\begin{figure}[t]
    \centering
    \includegraphics[width=\textwidth]{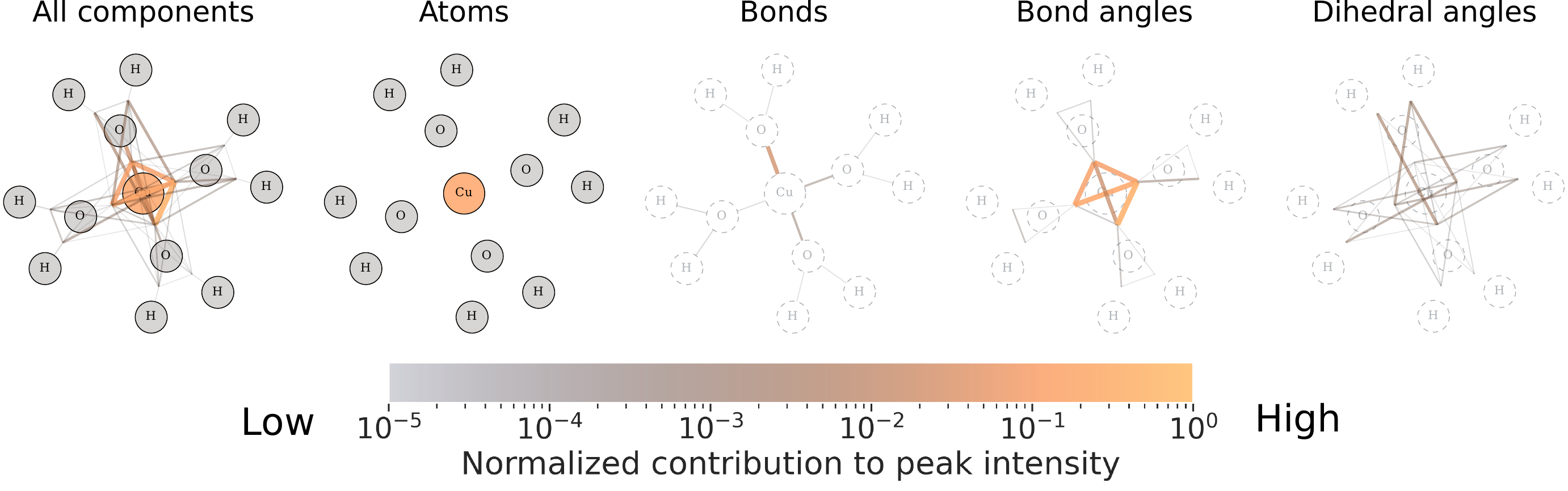}
    \caption{Graph visualization of the relative contributions of atomic, bond, bond angle, and dihedral angle components to spectral peak intensity for one fourfold-coordinated Cu(II) aqua configuration. The largest contributions are derived from the central copper and O-Cu-O bond angles, followed by dihedral angles and Cu-O bonds. Color and line thickness indicate the magnitude of the relative contributions. For the visualizations of the bond and angle components, the atoms are faintly outlined.}
    \label{fig:contrib-vis}
\end{figure}

Beyond efficiency and performance, ALIGNN-$d$ can be utilized for model interpretability. Specifically, the final output can be expressed as the sum of contributions from the individual graph components. In this way, relative contributions from the atoms, bonds, bond angles, and dihedral angles can be independently assessed. This is done by transforming the final embedding vectors (after the interaction layers shown in Fig.~\ref{fig:flow}) of the atoms, bonds, and angles into non-negative scalars, which are then summed to a scalar final output.

We trained this interpretable variant of ALIGNN-$d$ to predict the peak intensity $A_G$ of the infrared optical absorption spectral response of Cu(II) aqua complexes. The component-wise decomposition allows the contributions of the atomic, bond, and angular features from the model output to be directly visualized in graph format. This is illustrated in Fig.~\ref{fig:contrib-vis} for a specific aqua complex configuration, where we find that the peak intensity of optical response is primarily attributed to the central copper atom and O-Cu-O bond angles, followed by certain dihedral angles and Cu-O bonds. Other components have negligible contributions to peak intensity (note the logarithmic scale of the colorbar).

\begin{figure}[t]
    \centering
    \includegraphics[width=0.8\textwidth]{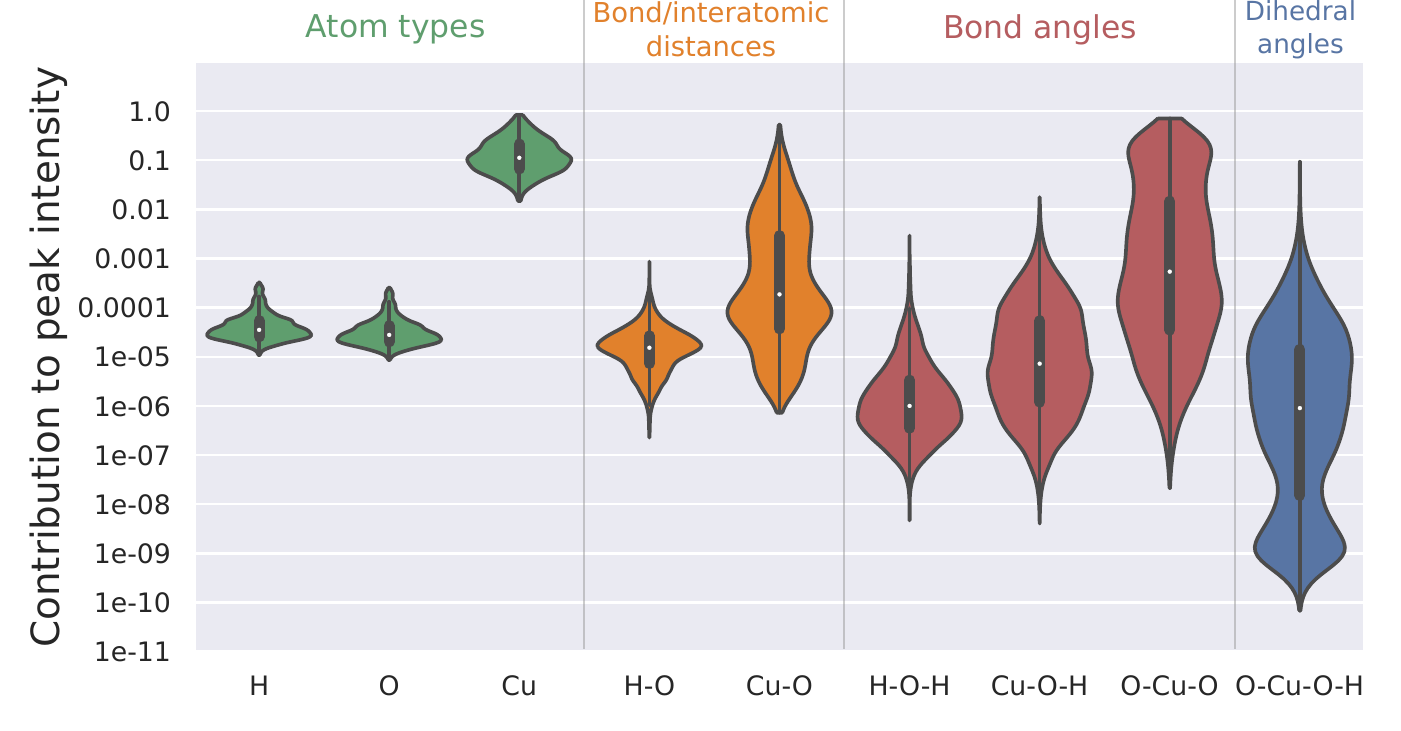}
    \caption{Normalized contributions to the peak intensity from individual graph components (atoms, bonds, bond angles, and dihedral angles). These contributions were obtained from the interpretable variant of ALIGNN$-d$.}
    \label{fig:contrib-hist}
\end{figure}

This same component decomposition procedure was then applied to all configurations of the aqua complexes. The results, presented in Fig.~\ref{fig:contrib-hist}, provide intuitive understanding of the physicochemical origins of the optical absorption response. As expected, the Cu atom features the highest contribution, consistent with the fact that changes in the $d$-shell electronic properties are largely responsible for the optical response. Nearby components that involve Cu atoms, including Cu--O bonds and O--Cu--O angles, also yield relatively higher contributions compared to others, such as water H--O--H angles. Overall, the bond angles contribute significantly to the model output, consistent with physical intuition that the angular information is critical for informing electronic properties in transition metal complexes. This points to the need for explicitly incorporating angular features in the graph representation.

Although contributions from dihedral angles are generally less significant than the O--Cu--O bond angles (which is expected given that they represent higher-order interactions), they remain relatively significant when compared to features such as H--O bonds and H--O--H angles. We therefore conclude that contributions from dihedral angles are important for resolving subtle structural differences, including those that result from weak geometric perturbations. In the spectral response, the dihedral contributions may thus be critical for interpreting and reproducing the fine structure.

\begin{figure}
    \centering
    \begin{subfigure}[c]{0.3\textwidth}
        \centering
        \includegraphics[width=\textwidth]{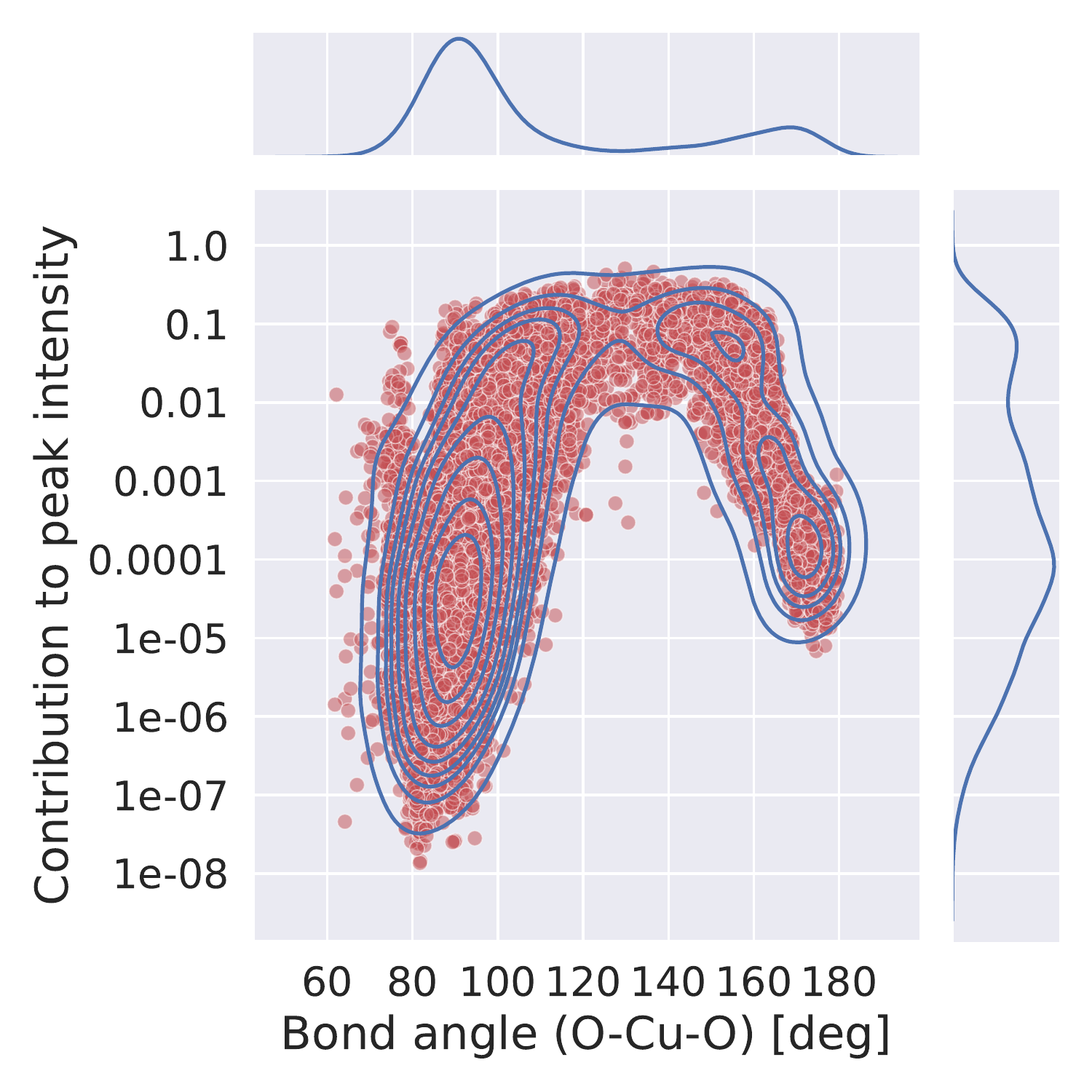}
        \caption{}
    \end{subfigure}
    \hfill
    \begin{subfigure}[c]{0.66\textwidth}
        \centering
        \includegraphics[width=\textwidth]{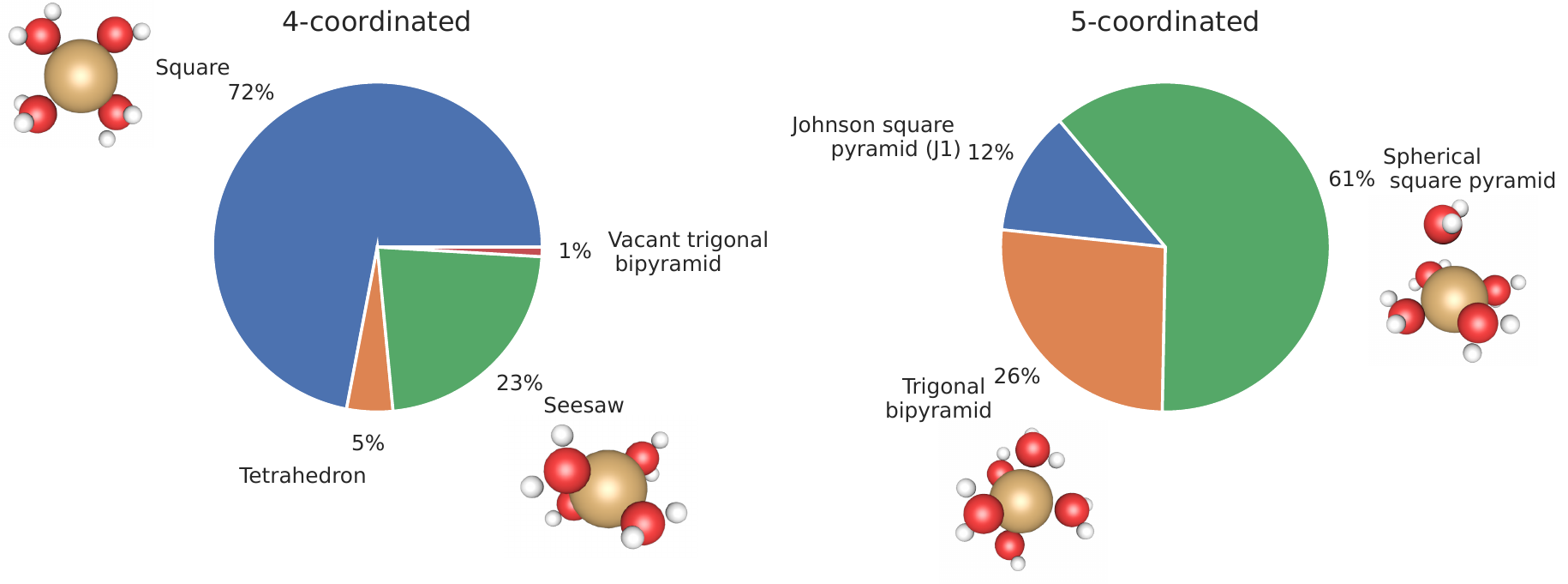}
        \caption{}
    \end{subfigure}
    \caption{(a) Relationship between O--Cu--O bond angles explored by the Cu(II) aqua complex during AIMD and their contributions to the peak intensity. (b) Distribution of closest-matching reference geometries based on the minimum CSM value, with AIMD snapshots illustrating the most commonly explored complexes. For this analysis, only fourfold- and fivefold-coordinated complexes are considered.}
    \label{fig:contrib-analysis}
\end{figure}

Additional information regarding the specific coupling between geometrical distortion and the infrared optical response can be obtained from further examination of the individual distributions in Fig.~\ref{fig:contrib-hist}. The Cu--O distance, O--Cu--O bond angle, and O--Cu--O--H dihedral distributions all display a degree of multimodality, suggesting that there are classes of geometries that contribute much more significantly to the optical response. To illustrate this further, we plot in Fig.~\ref{fig:contrib-analysis}a the relationship between peak intensity and bond angle for the specific case of the O--Cu--O bond angle distribution. The Cu(II) aqua complex is known to prefer octahedrally derived geometries, which is also common behavior across a range of other transition metal coordination complexes. Ideally, such geometries should exhibit angles of 90° and 180°. From Fig.~\ref{fig:contrib-analysis}a, we determine that configurations featuring these angles are minimal contributors to peak intensity. However, as the angles are even slightly perturbed ($>$ 90° or $<$ 180°), the peak intensity rapidly climbs several orders of magnitude. This is consistent with the physical understanding of the nominally symmetry-forbidden $d$-to-$d$ transitions that comprise the infrared optical response of the Cu(II) aqua complex, which require thermal distortion to remove the transition constraints.

Interestingly, it can be seen that the the highest contributions to peak intensity occur for bond angles in the range 120°-140°. These angles are not merely minor distortions, but rather represent new symmetries that are not octahedrally derived and have a much lower overall probability. To obtain further insight, we examined the symmetries explored during the AIMD simulations using the continuous shape measure (CSM) metric. Briefly, CSM provides a mathematically rigorous way to quantify similarity to reference geometric structures, from which closest matches to ideal symmetries can be assessed. 

Figure~\ref{fig:contrib-analysis}b shows the breakdown of instantaneous closest-matching CSM-derived geometries exhibited during our simulation trajectories, focusing on the fivefold- and fourfold-coordinated complexes that together represent the majority of all configurations. The CSM analysis confirms that the aqua complexes prefer to adopt the square- and pyramid-like configurations, which are octahedrally derived and dominated by 90° and 180° bond angles. The fourfold-coordinated complexes also feature the seesaw configuration, which is likewise octahedrally derived. However, a significant fraction of fivefold-coordinated complexes have the closest match to a trigonal bipyramid, which is geometrically distinct and features bond angles of 120°. These configurations are broadly reflective of the new symmetries in the 120°--140° region that contribute most strongly to the optical response according to Fig.~\ref{fig:contrib-analysis}a. We therefore propose that aqua complexes that temporarily adopt new symmetries while actively transitioning from their most common geometries are critical contributors to the optical absorption in the infrared regime. In aqua complexes, such distortions occur occasionally due to thermal/solvent-induced fluctuations. However, one may imagine engineering local environments in frozen or glassy systems to bias these preferences and artificially enhance the frequency of optically responsive configurations.

In summary, our graph representation ALIGNN-$d$ is shown to be memory-efficient and capable of capturing the full geometric information of atomic structures. While the original ALIGNN paper \cite{decost2021atomistic} focuses on general material property prediction for periodic, crystalline systems and small-scale molecules, our work instead focuses on disordered and distorted systems. We also show the unique interpretability of the ALIGNN-$d$ approach, which was applied to elucidate contributions of specific structural features of the Cu(II) aqua complexes to their infrared optical absorption spectral signatures. 

It is worth noting that aside from memory efficiency, the use of auxiliary line graphs does introduce additional computational burden; future study is therefore needed to thoroughly determine the computational cost and scalablity of ALIGNN and ALIGNN-$d$. We also point out that in contrast to paiNN~\cite{schutt2021equivariant}, ALIGNN-$d$ does not incorporate directional information and therefore cannot be used to predict tensorial properties or direction-specific properties. However, ALIGNN-$d$ and paiNN are not necessarily mutually exclusive formulations. In this regard, an interesting direction for future study is to combine angular encoding (with line graphs) and equivariant directional encoding. 

Finally, we point out that our framework is general and can be applied to other materials systems and properties. For instance, it could accelerate material design and selection of metal complexation with controlled shift in optical absorption for targeted optical and filtering properties. It could facilitate analysis of other spectroscopic responses in complex, disordered materials, which feature signatures that are often convoluted and difficult to interpret. It could also reveal features in the fine structure of spectra that might otherwise be overlooked. However, we caution that not all physical properties can be manifested as a simple summation of the graph components formulated in this work. In this regard, more elaborate interpretation methods may elucidate contributions of other specific structural or chemical features, opening the door to a broader range of applications. 

\section{Methods}

\subsection{Data preparation}
    \subsubsection{Molecular dynamics simulations}
    Solvated \ch{Cu^2+} ion was modeled using an ion in a cubic supercell with 48 water molecules at  the experimental density of liquid water at ambient conditions. We carried out Car-Parrinello molecular dynamics simulations using the Quantum ESPRESSO package \cite{QE}, with interatomic forces derived from density functional theory (DFT) and the Perdew-Burke-Ernzerhof (PBE) exchange-correlation functional \cite{PBE}. The interaction between valence electrons and ionic cores was represented by ultrasoft pseudopotentials \cite{UPF}; we used a plane-wave basis set with energy cutoffs of 30~Ry and 300~Ry for the electronic wavefunction and charge density, respectively. All dynamics were run in the \emph{NVT} ensemble at an elevated temperature of 380 K to correct for the overstructuring of liquid water at ambient temperatures with the PBE functional \cite{pbewater}. A time step of 8 a.u. were employed with an effective mass of 500 a.u. with hydrogen substituted with deuterium. The water was first equilibrated for 10 ps before the Cu$^{2+}$ ion was inserted into the system, after which the system was equilibrated for another 10 ps. This was followed by a 40 ps production run to extract the time-averaged properties of the system.  
    
    \subsubsection{Optical absorption spectroscopy simulation}
    Time dependent density functional theory (TDDFT)~\cite{tddft} within the Tamm–Dancoff approximation~\cite{TD} was used to obtain the optical excitation energies of the ion complexes. Specifically, 6,846 aqua complexes were extracted roughly uniformly from the AIMD trajectory. Each complex consists of the copper ion and the surrounding water molecules within a radial cutoff of 2.92 \AA, which corresponds to the first minimum of the Cu-O partial radial distribution function for the \ch{Cu^2+} oxidation state. All the optical calculations were carried out using the NWChem software package \cite{apra2020nwchem}. Here, an augmented cc-pVTZ basis was used for the copper ion while an augmented cc-pVDZ basis was used for the water molecules. Four examples of the aqua complexes and their corresponding TDDFT-calculated peaks in roughly the visible-infrared range are shown in Fig.~\ref{fig:clusters-and-peaks}. The aqua complexes data was randomly partitioned into a training set of 6,161 samples and a validation set of 685 samples.
    
    \subsubsection{High-throughput automation}
    To assist with the large number of copper complexes being studied, an AiiDA \cite{huber2020aiida} workflow was employed to standardize and provide consistency for all calculations. AiiDA records full provenance between all calculations and ensures a robust framework for generating, storing, and analyzing results.

    \subsubsection{Single-peak approximation}
    For GNN spectroscopy prediction, we focus on the TDDFT-calculated spectral peaks roughly in the visible-infrared range. Since the peak positions per complex tend to be close together, we approximated each complex's discrete peaks into a single unnormalized Gaussian curve. This approximation helps simplify the output format for training a structure-to-spectrum GNN model. For example, the original spectral peaks from TDDFT may be described by a set of tuples $(E_1, I_1), (E_2, I_2), ...$, where $E$ is the peak energy, and $I$ is the peak intensity. After the approximation, we can simply describe the spectral signature with an unnormalized Gaussian, parametrized by the mean $\mu_G$, the standard deviation $\sigma_G$, and the amplitude $A_G$. The single-peak approximation is further explained in Supplementary Information.
    
    \subsubsection{Continuous Shape Measure}
    The continuous shape measure (CSM) provides a mathematically rigorous, normalized measure of the deviation of a molecular fragment geometry from an ideal reference polyhedron. As a similarly metric, CSM is bounded between 0 and 100. Low CSM value indicates high similarity to the reference shape symmetry, and thus to a highly symmetric arrangement of water molecules around the copper ion. The CSM was computed using the SHAPE code~\cite{csm} for the entire standard set of polyhedral reference geometries for four- and five- coordination numbers. 

\subsection{ALIGNN-\textit{d} representation}
In the original ALIGNN formulation \cite{decost2021atomistic}, two graphs are used to encode one atomic structure: an original atomic graph $\G$ and its corresponding line graph $\LG$. The nodes and edges in $\G$ represent atoms and bonds, respectively. The nodes and edges in $\LG$, on the other hand, represent bonds and bond angles, respectively. Note that the edges in $\G$ and the nodes in $\LG$ are identical entities and share the same embedding during GNN operation. In this work, we extended the ALIGNN encoding to also explicitly represent dihedral angles. Different from the original ALIGNN paper, we encoded the atomic, bond, and angular features with only the minimally required information, namely the atom type $z$, the bond distance $d$, the bond angle $\alpha$, and the dihedral angle $\alpha'$. In other words, in $\G$, each node corresponds to a value of $z$, and each edge corresponds to a value of $d$. In $\LG$, each node also corresponds to a value of $d$, and each edge corresponds to a value of $\alpha$. Again, the edges in $\G$ are identical to the nodes in $\LG$ and $\LpG$. Finally, in $\LpG$, each edge representing a dihedral angle corresponds to a value of $\alpha'$. Information such as electronegativity, group number, bond type, and so on are not encoded.

We used Atomic Simulation Environment \cite{larsen2017atomic} and PyTorch Geometric \cite{fey2019fast} to construct the graph representations, and to calculate bond and dihedral angles.

\subsection{Model architectures}
Two different GNN model architectures were defined: one for $\G_{min}$ and $\G_{max}$; and one for $\G_{min} \cup \LG_{min}$ and $\G_{min} \cup \LpG_{min}$. We kept the two architectures as identical as we could in order to fairly evaluate and compare the model performances as a function of input graph representation. Both architectures consist of three parts: the initial encoding, the interaction operations, and the output layers (Fig.~\ref{fig:flow}).

In the initial encoding, the atom type, the bond distance, and the angular values are converted from scalars to feature vectors for subsequent neural network operations.
The atom type $z$ is transformed by an \verb|Embedding| layer (see PyTorch documentation \cite{paszke2019pytorch}).
The bond distance $d$ is expanded into a $D$-dimensional vector by the Radial Bessel basis functions, or RBF, proposed by Klicpera \textit{et al}. \cite{klicpera2020directional}.
The $n$th element in the vector is expressed as
\begin{equation} \label{eq:rbf}
    \text{RBF}_n(d) =
    \sqrt{\frac{2}{c}} \frac{\sin(\frac{n \pi}{c} d)}{d},
\end{equation}
where $n \in [1..D]$ and $c$ is the cutoff value. We also used RBF to expand angular information. However, the angular encoding treats bond angles $\alpha$ and dihedral angles $\alpha'$ differently, and encodes their values at different channels of the expanded feature vector. Further details regarding the angular encoding are described in Supplementary Information.

The interaction operations are also known as graph convolution, aggregation, or message-passing. Following the ALIGNN paper \cite{decost2021atomistic}, we also adopted the edge-gated graph convolution \cite{bresson2017residual, dwivedi2020benchmarking} for the interaction operations. The node features $\vec{h}^{l+1}_i$ of node $i$ at the $(l+1)$th layer is updated as
\begin{equation} \label{eq:node-update}
    \vec{h}^{l+1}_i =
    \vec{h}^l_i + 
    \mathrm{SiLU} \left(
        \mathrm{LayerNorm} \left(
            \vec{W}^l_s \vec{h}^l_i + 
            \sum_{j \in \mathcal{N}(i)} \hat{\vec{e}}^l_{ij} \odot \vec{W}^l_d \vec{h}^l_j
        \right)
    \right),
\end{equation}
where SiLU is the Sigmoid Linear Unit activation function \cite{elfwing2018sigmoid}; LayerNorm is the Layer Normalization operation \cite{ba2016layer}; $\vec{W}_s$ and $\vec{W}_d$ are weight matrices; the index $j$ denotes the neighbor node of node $i$; $\hat{\vec{e}}_{ij}$ is the edge gate vector for the edge from node $i$ to node $j$; and $\odot$ denotes element-wise multiplication. The edge gate $\hat{\vec{e}}^{l}_{ij}$ at the $l$th layer is defined as
\begin{equation} \label{eq:edge-gate}
    \hat{\vec{e}}^l_{ij} = 
    \frac
    {\sigma(\vec{e}^l_{ij})}
    {\sum_{j' \in \mathcal{N}(i)} \sigma(\vec{e}^l_{ij'}) + \epsilon},
\end{equation}
where $\sigma$ is the sigmoid function, $\vec{e}^l_{ij}$ is the original edge feature, and $\epsilon$ is a small constant for numerical stability.
The edge features $\vec{e}^l_{ij}$ is updated by
\begin{equation} \label{eq:edge-update}
    \vec{e}^{l+1}_{ij} =
    \vec{e}^l_{ij} + 
    \mathrm{SiLU} \left(
        \mathrm{LayerNorm} \left(
            \vec{W}^l_g \vec{z}^l_{ij}
        \right)
    \right),
\end{equation}
where $\vec{W}_g$ is a weight matrix, and $\vec{z}_{ij}$ is the concatenated vector from the node features $\vec{h}_i$, $\vec{h}_j$, and the edge features $\vec{e}_{ij}$:
\begin{equation}
    \vec{z}_{ij} = \vec{h}_i \oplus \vec{h}_j \oplus \vec{e}_{ij}.
\end{equation}
We applied the same edge-gated convolution scheme (Eq.~\ref{eq:node-update}--\ref{eq:edge-update}) to operate on both the atomic graph $\G$ and the line graphs $\LG$, $\LpG$. In the case of $\G$, the edge-gated convolution updates nodes that represent atoms, and edges that represent bonds, while exchaning information between the two, hence the term \emph{atom-bond interaction} shown in Fig.~\ref{fig:flow}. In the case of $\LG$, the convolution updates nodes that represent bonds, and edges that represent angles, hence the term \emph{bond-angle interaction} shown in Fig.~\ref{fig:flow}. Note that by iteratively applying the convolution operation on $\G$ and $\LG$, the angular information stored in $\LG$ can propagate to $\G$. Due to the nature of the edge-gated convolution, all the feature/embedding vectors for atoms, bonds, and angles during the interaction layers have the same length, or the same number of channels $D$.

Lastly, the final output layers pool (by summation) the node features of $\G$ and transform the pooled embedding into an output vector, which is a three-dimentional vector consisted of the parameters of an unnormalized Gaussian curve $\mu_G$, $\sigma_G$, and $A_G$. The two \verb|Linear| layers have the output lengths of 64 and 3, respectively. For the interpretable variant of the model, the final output layers are replaced by a \verb|Linear| layer transforming the input vectors into scalars, followed by the softplus activation $\log(1+\exp(\cdot))$ and global summation. This operation effectively transforms each embedding vector into a non-negative scalar before summing all the scalars into a positive scalar final output. Therefore, these component scalars can be interpreted as the atomic, bond, and angular contributions to the final output.

The model parameters for both architectures in Fig.~\ref{fig:flow} are the same, and listed in Table~\ref{tab:model-params}.

\begin{table}
    \centering
    \caption{Model parameters}
    \begin{tabular}{l l l}
    \hline
    Name & Notation & Value \\
    \hline
    Number of interaction layers            & $L$               & 6                     \\
    RBF cutoff (bond distance)              & $c_d$             & 6.0 \si{\angstrom}    \\
    RBF cutoff (cosine and sine angles)     & $c_{\alpha}$      & 2.0                   \\
    Number of channels                      & $D$               & 64                    \\
    \hline
    \end{tabular}
    \label{tab:model-params}
\end{table}

\subsection{Model training}
We used PyTorch Geometric \cite{fey2019fast} to develop the GNN models. Note that although two model architectures were defined, a total of four GNN models were trained due to the four different graph representations studied in this work. Nonetheless, these models have the same parameters (Table~\ref{tab:model-params}). Similar to the ALIGNN paper \cite{decost2021atomistic}, we trained each model using the Adam optimizer \cite{kingma2014adam} and the 1cycle scheduler \cite{smith2019super}. Each training was carried out in PyTorch \cite{paszke2019pytorch} and PyTorch Geometric \cite{fey2019fast} on a NVIDIA V100 (Volta) GPU, and repeated eight times with randomly initialized weights for statistical robustness. The mean squared error (MSE) was used as the loss during training.
The training parameters are the same for each training (Table~\ref{tab:train-params}). All other parameters, if unspecified in this work, default to values per PyTorch 1.8.1 and PyTorch Geometric 1.7.2.

\begin{table}
    \centering
    \caption{Training parameters}
    \begin{tabular}{l l l}
    \hline
    Name & Notation & Value \\
    \hline
    Batch size                          & $M$                       & 128                   \\
    Number of epochs                    & $N_\mathrm{ep}$           & 1000                  \\
    Initial learning rate               & $\eta_{\mathrm{init}}$    & 0.0001                \\
    Maximum learning rate  (1cycle)     & $\eta_{\mathrm{max}}$     & 0.001                 \\
    First moment coefficient for Adam   & $\beta_1$                 & 0.9                   \\
    Second moment coefficient for Adam  & $\beta_2$                 & 0.999                 \\
    \hline
    \end{tabular}
    \label{tab:train-params}
\end{table}

\section*{Acknowledgements}
The authors are partially supported by the Laboratory Directed Research and Development (LDRD) program (20-SI-004) at Lawrence Livermore National Laboratory. This work was performed under the auspices of the US Department of Energy by Lawrence Livermore National Laboratory under contract No. DE-AC52-07NA27344.

\section*{Author Contributions}
T. A. Pham, S. R. Qiu, X. Chen, and B. C. Wood supervised the research.
B. C. Wood computed the MD trajectory of the solvated copper ion.
N. Keilbart developed the automated Aiida workflow for TDDFT calculations, with assistance from S. Weitzner.
T. Hsu performed the TDDFT calculations, developed the ALIGNN-d representation, and trained the GNN models.
T. Hsu, T. A. Pham, and B. C. Wood wrote the manuscript with inputs from all authors.

\section*{Competing interests}
On behalf of all authors, the corresponding author states that there is no conflict of interest.

\section*{Data Availability}
All data required to reproduce this work can be requested by contacting the corresponding author.


\printbibliography
\renewcommand{\thepage}{S\arabic{page}} 
\renewcommand{\thesection}{S\arabic{section}}  
\renewcommand{\thetable}{S\arabic{table}}  
\renewcommand{\thefigure}{S\arabic{figure}}
\setcounter{figure}{0}

\section*{Supplementary Information}

\subsection*{Shape analysis}
\begin{figure}
    \centering
    \begin{subfigure}[b]{0.8\textwidth}
        \centering
        \includegraphics[width=\textwidth]{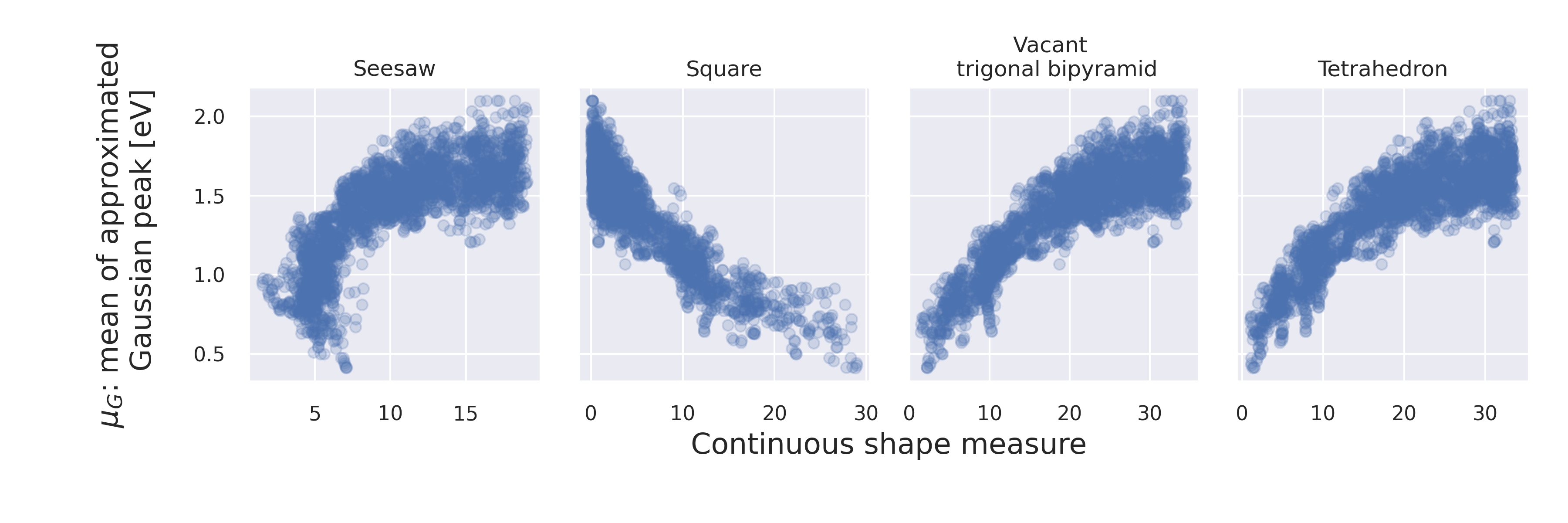}
        \caption{4 oxygens in each cluster}
    \end{subfigure}
    \begin{subfigure}[b]{0.8\textwidth}
        \centering
        \includegraphics[width=\textwidth]{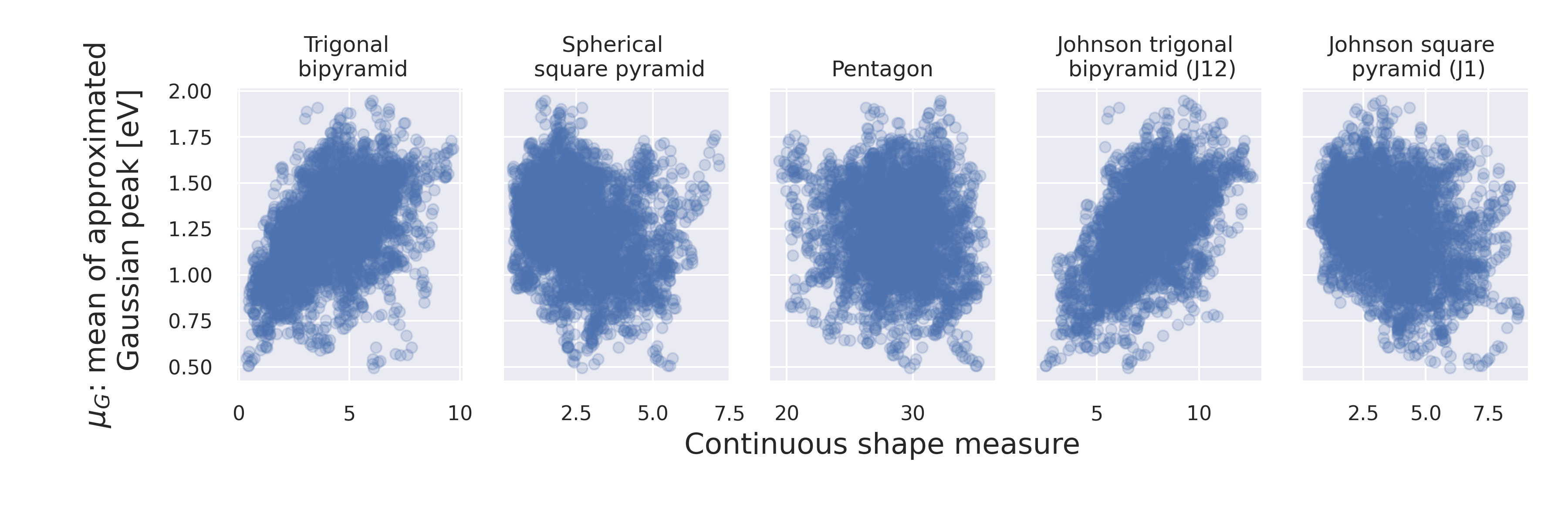}
        \caption{5 oxygens in each cluster}
    \end{subfigure}
    \caption{A solvated copper cluster’s shape information is correlated with its spectral information, as shown in these scatter plots, where each scatter point represents a cluster structure. The shape quantity (x-axis) is continuous shape measure (CSM), a measure of deviation from an ideal reference polyhedron (indicated by subplot title), whereas the spectral quantity (y-axis) is the mean of the fitted Gaussian peak (see Fig.~\ref{fig:single_peak_approx}). This shape-spectrum correlation is especially strong for clusters containing 4 oxygens or water molecules (a), but less so for clusters containing 5 oxygens (b). Nonetheless, in (b), there is some observable correlation between two CSM quantities (trigonal bipyramid and Johnson trigonal bipyramid) and the spectral quantity.}
    \label{fig:cshm_corr}
\end{figure}

We calculated the CSM \cite{pinsky1998continuous, casanova2004minimal, cirera2006shape} for the shapes formed by the central copper atom and the coordinated oxygen atoms from the copper cluster data. In the case of 4 oxygens surrounding the copper atom, the ideal reference shapes are square, tetrahedron, sawhorse, and axially vacant trigonal bipyramid.
In the case of 5 oxygens, the ideal reference shapes are pentagon, vacant octahedron (or the $J_1$ Johnson solid), trigonal bipyramid, square pyramid, and the Johnson trigonal bipyramid ($J_{12}$).
For each reference shape, the CSM is plotted against the mean of the approximated Gaussian peak $\mu_G$, as shown in Fig.~\ref{fig:cshm_corr}.
Significant correlation between CSM and $\mu_G$ can be clearly observed with respect to certain reference shapes.
This result suggests that the spectral signature of the solvated copper clusters is a geometry-sensitive property.
Note that the CSM quantities here do not account for the hydrogen atoms in the clusters.
Also, clusters containing 6 oxygens are rare and thus were omitted.

\newpage
\subsection*{Single-peak approximation}
The single-peak approximation of the discrete spectral lines is based on peak broadening, followed by a simple least-square fitting of a single unnormalized Guassian curve.
The broadening step is equivalent to kernel density estimation
\begin{equation}
    \hat{f}(x) = \sum_{i=1}^{N} K(x - x_i),
\end{equation}
where $\hat{f}(x)$ is the broadened function or spectrum over the energy values $x$, $N$ is the number of discrete peaks, and $K$ is the kernel function.
We used a Gaussian kernel with a standard deviation of 0.2 eV.
The least-square fitting was implemented using SciPy \cite{2020SciPy-NMeth}. The fitted Gaussian curve 
\begin{equation}
    g(x) = A_G \
        \frac{1}{\sigma_G \sqrt{2 \pi}} \
        \exp \frac{-1}{2} \left(\frac{x - \mu_G}{\sigma_G} \right)^2
\end{equation}
is parametrized by the mean $\mu_G$, the standard deviation $\sigma_G$, and the amplitude $A_G$. 

\begin{figure}
    \centering
    \includegraphics[width=0.7\textwidth]{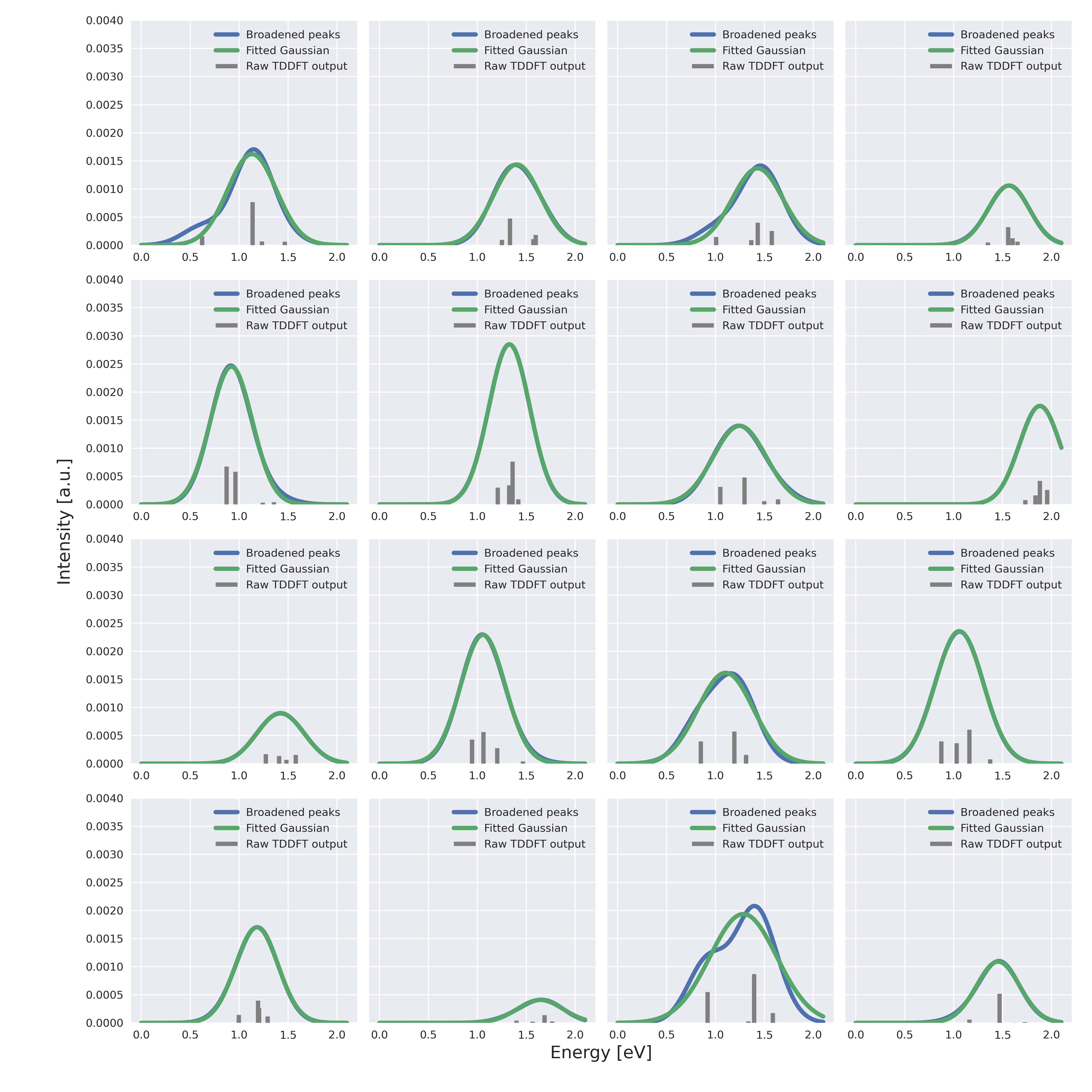}
    \caption{The TDDFT-calculated discrete spectral peaks (vertical, gray lines) in roughly the visible-infrared range tend to be close together and were approximated to be a single Gaussian curve. The approximation is based on first broadening the discrete peaks with a Gaussian kernel and then fitting the broadened peaks (blue curves) with a Gaussian function (green curves). 16 randomly sampled spectral signatures and their corresponding approximations are shown.}
    \label{fig:single_peak_approx}
\end{figure}

\newpage
\subsection*{Sampled predicted and target spectral peaks}
\begin{figure}
    \centering
    \includegraphics[width=0.8\textwidth]{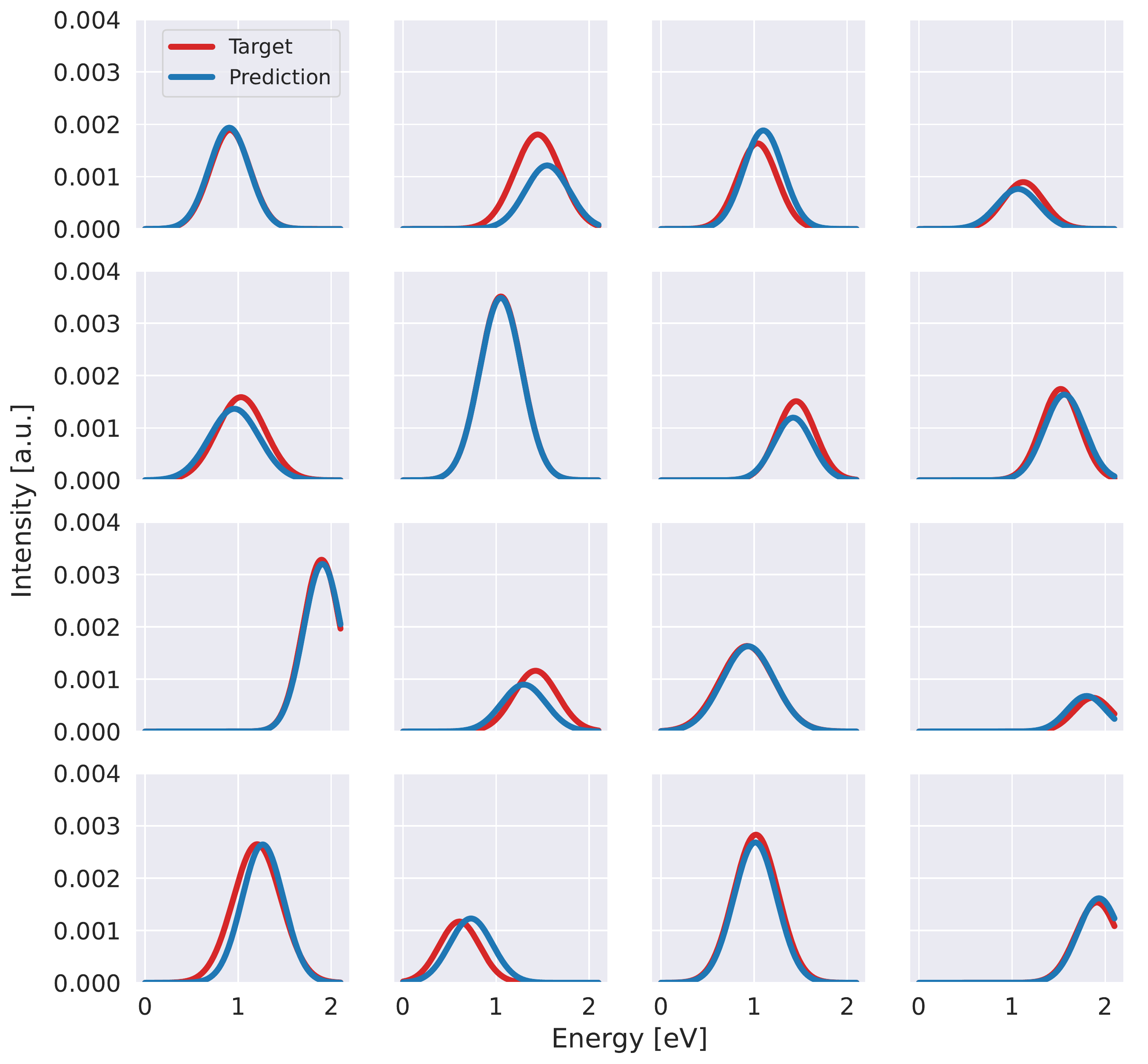}
    \caption{16 randomly sampled GNN-predicted spectral peaks (based on the ALIGNN-$d$ representation) and the corresponding target peaks. These predictions were made on the validation dataset.}
    \label{fig:pred-vs-true}
\end{figure}

Based on the ALIGNN-$d$ representation, 16 randomly sampled GNN-predicted spectral peaks and their corresponding target spectral peaks are shown in Fig~\ref{fig:pred-vs-true}. These results further verify that the trained GNN provides accurate spectroscopic prediction.

\newpage
\subsection*{Angular encoding}
The angular encoding treats bond angles $\alpha$ and dihedral angles $\alpha'$ as two different types of angles, and encodes their values at different channels of the expanded feature vectors $\vec{e}$, which correspond to the edges of line graphs.
The bond angle is encoded in the first half of $\vec{e}$, and the dihedral angle is encoded in the second half.
While the bond angle ranges from 0° to 180°, the dihedral (torsion) angle ranges from 0° to 360°.
The use of trigonometric functions (sine and cosine) informs the model the periodic nature of the dihedral angle, i.e., there is little physical difference between 1° and 359°.
It is also necessary to use both sine and cosine functions to retain the full dihedral angle information.
Therefore, the encoding of the dihedral angle is further divided into the cosine and sine components, each occupying a quarter of the channels of the feature vector. The unoccupied parts of the feature vector are initialized with zeros. Lastly, since the sine and cosine values range from -1 to 1, these values are added by 1 prior to the RBF expansion with a cutoff value of $c = 2$.

\begin{figure}
    \centering
    \includegraphics[width=0.8\textwidth]{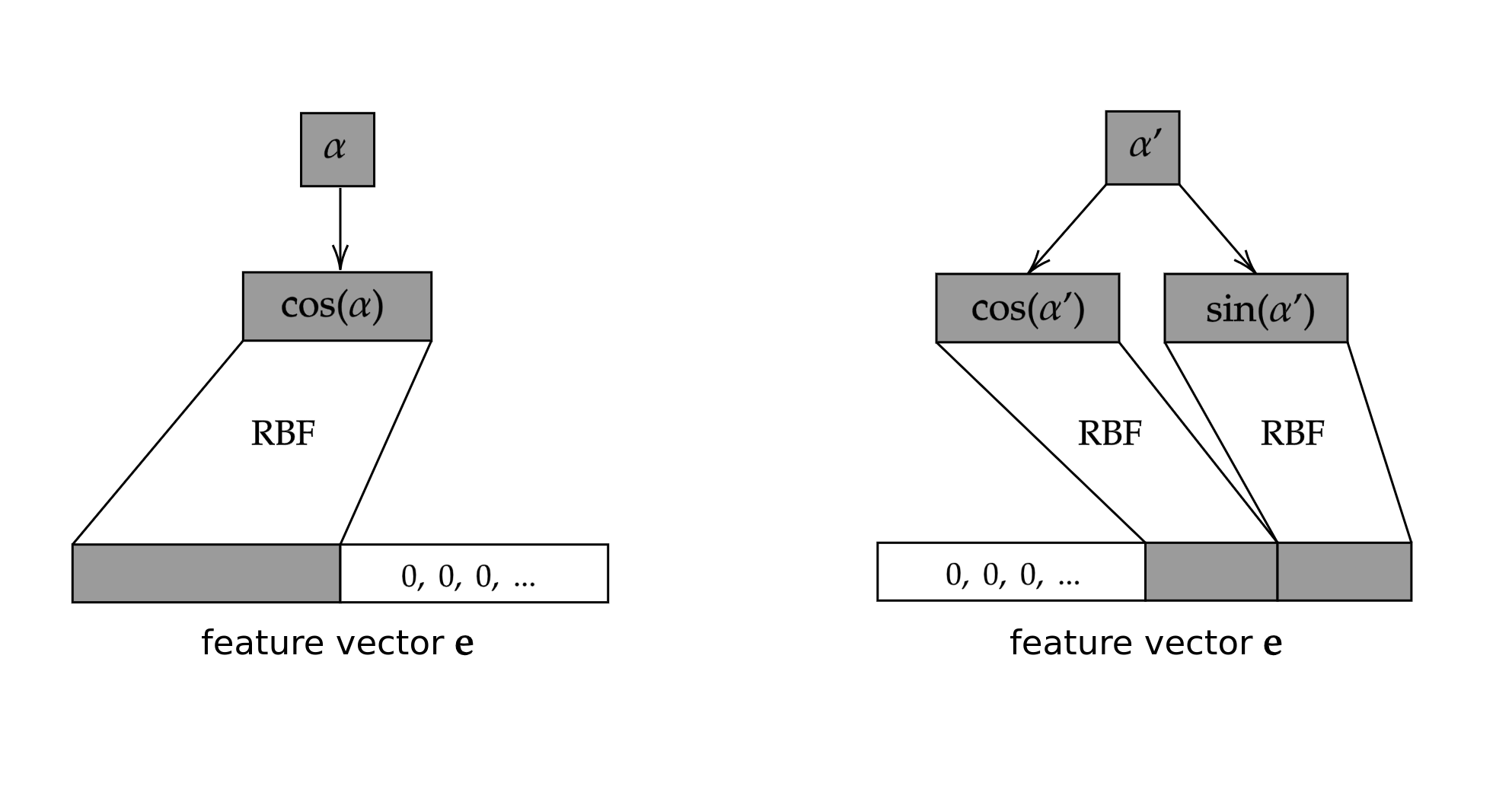}
    \caption{Angular encoding for bond angles $\alpha$ and dihedral angles $\alpha'$. The angular values, in the form of sine and cosine functions, are expanded into feature vectors by RBF.}
    \label{fig:encode-ang}
\end{figure}

\end{document}